\documentclass[letterpaper, 10 pt, conference]{ieeeconf}  % Comment this line out if you need a4paper

\IEEEoverridecommandlockouts                              % This command is only needed if 
\usepackage{graphics} % for pdf, bitmapped graphics files
\usepackage{epsfig} % for postscript graphics files
\usepackage{mathptmx} % assumes new font selection scheme installed

\usepackage{times} % assumes new font selection scheme installed
\usepackage{amsmath} % assumes amsmath package installed
\usepackage{amssymb}  % assumes amsmath package installed
\usepackage{float}
\usepackage{array}
\usepackage{multirow}
\usepackage{xcolor}
\usepackage{algorithm}
\usepackage{algpseudocode}
\usepackage{amsmath}
\usepackage{amssymb}
\usepackage{soul}
\usepackage{booktabs}
\usepackage{multirow}
\usepackage{makecell}
\usepackage{array}
\usepackage{booktabs}

\usepackage{caption}
\usepackage{booktabs}
\usepackage{multirow}
\usepackage{makecell}
\renewcommand{\arraystretch}{1.15}

\definecolor{multi_norm_voxel}{rgb}{0.0, 0.8, 0.8}

\definecolor{multi_norm}{rgb}{0.3, 0.6, 1.0}

\algrenewcommand\algorithmicrequire{\textbf{Input:}}
\algrenewcommand\algorithmicensure{\textbf{Output:}}

\newcommand{\legendbox}[1]{%
\raisebox{0.2ex}{\textcolor{#1}{\rule{1ex}{1ex}}}%
}

\title{\LARGE \bf
Ground Sample Distance Constrained $k$-Coverage Viewpoint Planning via Set Multi-Cover Formulation
}

\author{Waleed Bin Khalid and Chul Min Yeum%
\thanks{Waleed Bin Khalid is a Ph.D. student in Computer
Science at the Luddy School of Informatics, Computing,
and Engineering, Indiana University Bloomington, IN, USA. Chul Min Yeum is an Associate Professor of Civil and Environmental Engineering,
University of Waterloo, ON, Canada.}%
}

\begin{document}

% Add author names and affiliations before \maketitle for the arXiv version.
\maketitle
\thispagestyle{empty}
\pagestyle{empty}

\begin{abstract}
Drone-based 3D reconstruction and inspection require viewpoints that satisfy image-quality constraints while observing each surface region multiple times, which overlap-based planning heuristics do not guarantee. We study ground sample distance (GSD)-constrained $k$-coverage viewpoint selection for structures represented by surface voxels. Candidate viewpoints are generated along outward surface directions, screened with distance, viewing-angle, field-of-view, and occlusion tests, and augmented for voxels with insufficient visibility. Selection is then posed as set multi-cover with per-voxel coverage demands. We propose a dynamically weighted greedy heuristic that prioritizes remaining coverage deficits and scarce viewing alternatives, and compare it with sequential overlap-based pruning and an integer linear programming (ILP) formulation on the same candidate pools. Across two structures, four voxel resolutions, and three coverage requirements, the proposed approach achieves full or near-full $k$-coverage with compact viewpoint sets, using only 3.00--16.39\% more viewpoints than ILP while offering an interpretable selection process based on coverage deficits and viewpoint availability.
\end{abstract}

\section{Introduction}
Autonomous aerial inspection and 3D reconstruction require a set of camera viewpoints that observe all surface regions and satisfy image-quality constraints governing reconstruction fidelity. In such applications, image resolution is commonly expressed through ground sampling distance (GSD), which directly determines the recoverable surface detail in the reconstructed model \cite{wilkinson2014elements,Schenk2005,Kraus2011}. 

Photogrammetric flight planning commonly recommends $60$--$80\%$ image overlap to support reliable reconstruction \cite{wilkinson2014elements,Forstner2016}. Under idealized strip-imaging assumptions, a forward overlap $\rho$ implies that a surface point appears in approximately $1/(1-\rho)$ consecutive images, motivating multi-view redundancy. However, overlap is a global acquisition heuristic and does not directly enforce that each surface region is observed a prescribed number of times under sensing constraints, particularly in the presence of self-occlusions and complex geometry.

In this study, we address the viewpoint generation and selection problem for aerial inspection of structures with known geometry (in a world coordinate frame). Specifically, we aim to generate a compact set of camera viewpoints, each defined by a waypoint and a camera orientation, that satisfies prescribed GSD and $k$-coverage requirements. The GSD constraint enforces a spatial-resolution bound under the adopted sensing model, while the $k$-coverage requirement seeks observations of each surface region from at least $k$ distinct viewpoints to support multi-view observation.

We formulate inspection viewpoint planning as a \emph{GSD-constrained $k$-coverage} problem over a voxelized surface representation. Surface voxels represent discrete observable regions, and candidate viewpoints are generated to satisfy geometric constraints derived from a prescribed GSD bound and maximum oblique viewing angle. Each candidate viewpoint observes a subset of surface voxels, yielding a set multi-cover formulation that seeks at least $k$ observations per voxel, subject to the visibility available in the candidate set.

We construct a sensing-compliant candidate set and select a smaller subset of viewpoints using a globally informed greedy $k$-coverage heuristic, evaluated against an integer linear programming (ILP) reference formulation over the same candidate set. Viewpoint selection is explicitly decoupled from downstream tour generation, allowing discrete coverage requirements to be evaluated independently of trajectory optimization. The resulting framework enables explicit evaluation of sensing-constrained multi-view coverage under the adopted geometric sensing model.

\begin{figure*}[t]
    \centering
    \includegraphics[width=\linewidth]{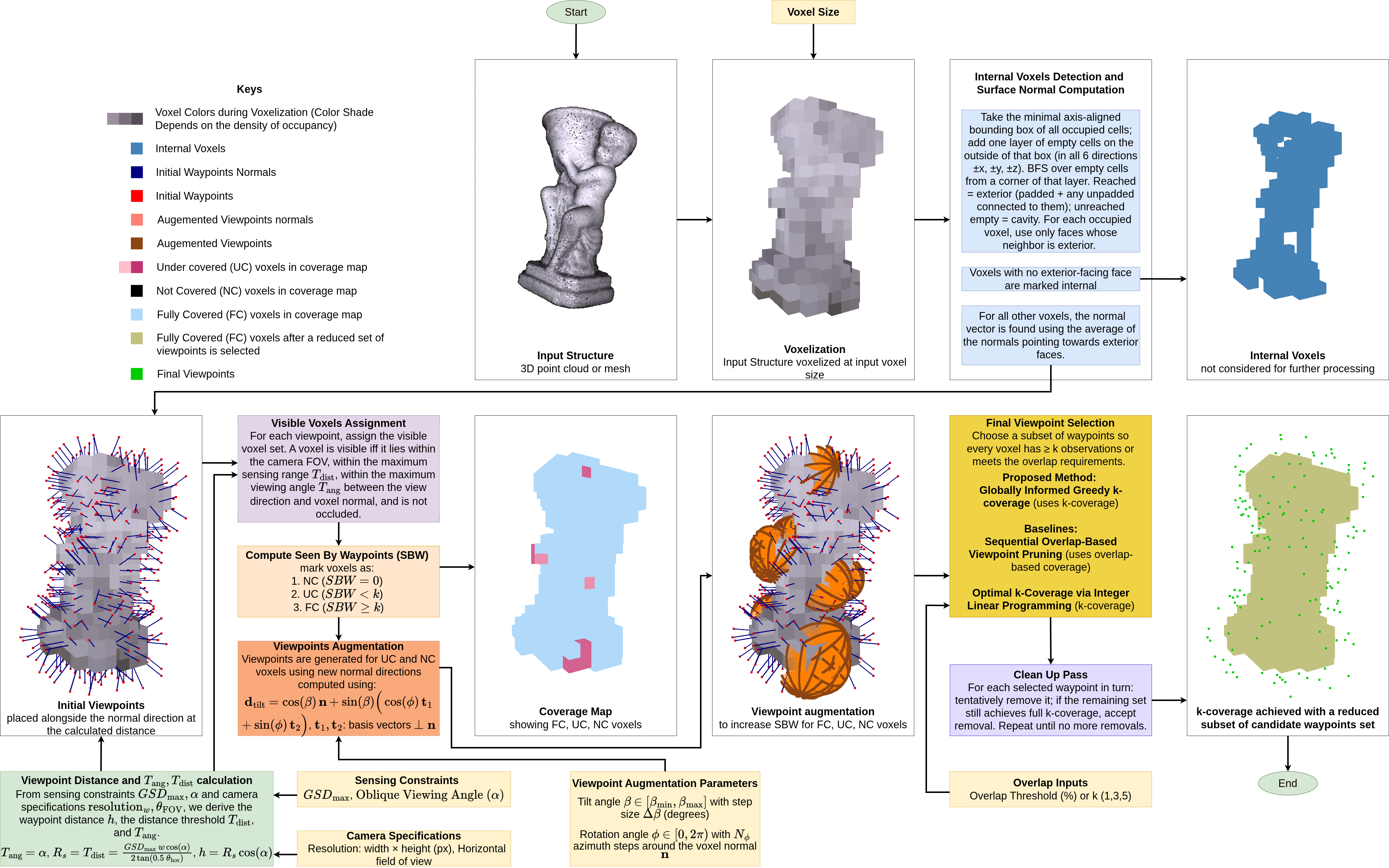}
    \caption{Overview of the proposed GSD-constrained $k$-coverage pipeline, from voxelization and sensing-compliant candidate generation to visibility assignment, augmentation, and final viewpoint selection.}
    \label{fig:block_diag}
\end{figure*}

\vspace{-1.5mm}

\section{Related Work}
\label{sec:related}

Coverage path planning (CPP) ensures a robot observes or traverses all regions of an environment. Early work established complete coverage through geometric decomposition and topological reasoning \cite{acar2002sensor}, with surveys formalizing CPP taxonomies and coverage--efficiency trade-offs \cite{choset2001coverage,galceran2013survey,tan2021comprehensive}. These approaches provide spatial coverage guarantees but primarily reason about traversal completeness rather than image (sensing) quality or multi-view observations.

Modern inspection planners extend CPP to sensing-driven tasks by jointly selecting viewpoints and motion plans under geometric and robotic constraints \cite{faigl2011sensor}. Aerial inspection planners generate candidates through iterative or adaptive sampling to reconstruct complex structures \cite{bircher2015structural,almadhoun2018coverage}, while recent works incorporate sensing constraints such as field-of-view limits, incidence angle, and overlap heuristics \cite{lee2022ceo,liang2025optimized}. Reconstruction-oriented planners further improve imaging quality through continuous trajectory optimization or topology-aware planning \cite{zhang2021continuous,shang2023topology}.

Several systems have been proposed that couple viewpoint generation with trajectory optimization \cite{lu2023tmstc,tripicchio2023smooth,mitra2025online}. While such integrated approaches consider both sensing and motion objectives, our work focuses specifically on viewpoint selection, allowing per-surface-region coverage requirements to be examined independently of trajectory optimization.

Sequential redundancy reduction has been used in viewpoint planning, where candidate viewpoints are discarded if their visibility regions overlap substantially with previously selected views \cite{bircher2015structural,almadhoun2018coverage}. However, overlap-based pruning does not, by itself, enforce a prescribed number of observations for each surface region, and its outcome can depend on viewpoint ordering and overlap thresholds.

Recent work has formulated viewpoint planning as a combinatorial set-cover problem. Adaptive spatial subdivision combined with ILP optimization has been used to obtain minimum-cardinality viewpoint sets with complete feature coverage \cite{spensieri2025unified,ramesh2024approximate}. Related multi-cover formulations have also been studied in geometric coverage problems, demonstrating the use of explicit redundancy constraints in combinatorial optimization. These formulations establish mathematical approaches to coverage selection. Our work builds on this perspective by incorporating GSD-constrained candidate generation, coverage-driven viewpoint augmentation, and per-surface-region multi-cover requirements.

Voxel representations are widely used to simplify visibility reasoning in complex environments. Voxelized planners generate viewpoints around clustered surface regions prior to trajectory optimization \cite{hagmanns2024path}, and enriched voxel maps support planning in dynamic environments \cite{golpour20254d}.

Existing work has addressed geometric coverage, sensing-constrained viewpoint generation, and combinatorial viewpoint optimization through different formulations. Building on these approaches, this work considers their integration for GSD-constrained $k$-coverage over voxelized surfaces. Specifically, we combine sensing-aware candidate generation, targeted augmentation of undercovered surface regions, and set multi-cover optimization to select compact viewpoint sets satisfying prescribed coverage requirements when feasible.

\section{Viewpoint Generation}

\subsection{Ground Sample Distance (GSD) Formulation}

In the general case where the sensor images the surface at an oblique angle, the GSD is given by
\begin{equation}
\text{GSD} =
\frac{R_s \times 2\tan(0.5\theta_{FOV})}
{w \times \cos(\alpha)}
\label{eq:GSD as a function of horizontal fov}
\end{equation}
where $R_s = \sqrt{d^2 + h^2}$ is the slant distance between the sensor and the surface, $d$ is the lateral offset relative to the surface normal, $h$ is the perpendicular distance from the sensor to the surface, $w$ is the horizontal number of pixels, $\theta_{FOV}$ is the horizontal field of view of the camera, and $\alpha$ is the oblique viewing angle between the camera viewing direction and the surface normal. For our algorithm, we require that the GSD evaluated at each visible voxel center does not exceed a prescribed $\text{GSD}_{\max}$ under the adopted geometric sensing model. To enforce this constraint, we fix a maximum $\alpha$ value and compute the corresponding slant range $R_s$ using $\text{GSD}_{\max}$ and $\alpha$. Viewpoints are then placed at the perpendicular component of this slant range, $h = R_s \cos(\alpha)$, along the outward surface normal. Under this placement, the nominal nadir GSD becomes $\text{GSD}_{\text{set}} = \text{GSD}_{\max}\cos^2(\alpha)$.

\subsection{Voxelization}

Given a point cloud or mesh, we select a voxel size $s$ and construct a uniform 3D grid. Each voxel $\mathbf{v}$ has a grid index $(i, j, k)$, and its center is computed as $\mathbf{c}_{ijk} = \mathbf{o}_w + s \cdot (i + 0.5, \; j + 0.5, \; k + 0.5)$, where $\mathbf{o}_w$ is the origin.

\subsection{Labeling Exterior and Interior Space}

Some empty cells lie within sealed cavities unreachable from outside. To distinguish exterior space from interior cavities, the grid is padded with a layer of empty cells on all sides, and a flood-fill is performed from a corner of this boundary through empty cells only. Reached cells are labeled \emph{exterior}; unreached cells are classified as \emph{interior cavities}.

\subsection{Computing Normals}

Each surface voxel requires an outward-pointing normal to place a camera at a physically meaningful viewing direction. A direct option is to average the normals of mesh or point-cloud samples within the voxel, but this is input-dependent and sensitive to within-voxel sampling density, where sparse or uneven samples can yield unstable normals.

We therefore compute voxel normals from \emph{exterior-facing faces} only, using the exterior labeling from the previous step. For each occupied voxel, we examine its 6-connected neighbors. A face is \emph{exterior-facing} if the neighbor cell is empty and labeled exterior. If a voxel has no exterior-facing face, it is classified as internal (e.g., fully surrounded by occupied cells or facing only cavities) and is not assigned a normal. For all others, the voxel normal is the normalized sum of the corresponding face normals:
\[
\mathbf{n}_{\text{voxel}} =
\frac{\sum_{\text{exterior-facing faces}} \mathbf{n}_{\text{face}}}
{\left\|\sum_{\text{exterior-facing faces}} \mathbf{n}_{\text{face}}\right\|}
\]

If this sum is negligible due to opposing exterior faces on a thin wall or edge, the voxel generates multiple viewing directions. In such cases, each exterior-facing face normal is stored as a separate \emph{multi-normal} record, generating one candidate viewpoint per direction. Because only exterior-facing faces are considered, internal voxels and inward-pointing normals are not produced, resulting in outward-facing directions for surface voxels.

\subsection{Generating Initial Viewpoints}

For each surface voxel, an initial candidate viewpoint is placed at the GSD-derived distance $h$ along the normal direction using $\mathbf{p} = \mathbf{c}_{\text{voxel}} + h \cdot \mathbf{n}_{\text{voxel}}$. The viewpoint direction vector is defined as $\mathbf{o} = -\mathbf{n}_{\text{voxel}}$. The camera orientation at each viewpoint is defined by aligning its optical axis with $\mathbf{o}$, so that the sensor directly faces the generating surface element. For multi-normal voxels, one viewpoint is generated per normal direction.

\subsection{Removing Blocked Viewpoints}

Some viewpoints may have an occluded line of sight to their generating voxel. We apply a ray pre-filter in which for each viewpoint at $\mathbf{p}$, a ray is cast along its direction vector, $\mathbf{o}$, towards the target voxel and discarded if the first occupied intersection lies closer than the expected distance. A tolerance $\tau = s\sqrt{3}/2$ compensates for voxel surface intersections.

\subsection{Assigning Visible Voxels to a Viewpoint and Identifying Undercovered Voxels}

The GSD-derived viewpoint placement establishes sensing constraints under the adopted geometric model. Since $\cos^2(\alpha) \leq 1$, the nominal GSD setpoint satisfies $\mathrm{GSD}_{\mathrm{set}} \leq \mathrm{GSD}_{\max}$. Candidate visibility is subsequently evaluated using distance and viewing-angle thresholds, defined as $T_{\mathrm{dist}} = R_s$ and $T_{\mathrm{ang}} = \alpha$, respectively. These constraints are applied to voxel-center observations and provide a discrete approximation of sensing compliance rather than a guarantee over the entire voxel surface.

Each viewpoint, initially generated from a single voxel, may observe multiple voxels simultaneously. A voxel with center $\mathbf{c}_v$ and normal vector $\mathbf{n}_{v}$ is considered visible from viewpoint $\mathbf{p}$ if it satisfies distance, deviation, FOV, and occlusion constraints \cite{lee2022ceo}. The distance constraint requires $\|\mathbf{c}_v - \mathbf{p}\| \le T_{\text{dist}}$ and the deviation constraint requires $\arccos(-\mathbf{o} \cdot \mathbf{n}_v) \le T_{\text{ang}}$. The FOV constraint requires $\arccos(\hat{\mathbf{o}} \cdot \hat{\mathbf{q}}) \le 0.5\theta_{FOV}$, where $\mathbf{q} = \mathbf{c}_v - \mathbf{p}$. Finally, occlusions are checked using the same ray test employed for blocked-viewpoint removal, retaining only unobstructed visibility relations under the adopted model.

Each surface voxel is assigned a \emph{seen-by-waypoints} (SBW) count, defined as $\mathrm{SBW}(v)=\sum_{\mathbf{p}_j\in P}\mathbb{I}[v\in V_j]$, where $P$ denotes the generated candidate viewpoint set, $V_j$ is the set of surface voxels visible from viewpoint $\mathbf{p}_j$, and $\mathbb{I}[\cdot]$ is the indicator function. Voxels with $\mathrm{SBW}(v)\geq k$ have sufficient candidate visibility to satisfy the requested $k$-coverage, while $0<\mathrm{SBW}(v)<k$ indicates insufficient candidate visibility and $\mathrm{SBW}(v)=0$ indicates no candidate visibility. Voxels with $\mathrm{SBW}(v)<k$ are therefore identified for targeted viewpoint augmentation.

\subsection{Viewpoint Augmentation}

For voxels with $\mathrm{SBW}(v) < k$, additional candidate viewpoints are generated to address coverage deficiencies. New viewing directions are sampled within a cone centered on the voxel's outward normal, introducing alternative oblique viewpoints that may improve candidate visibility in hard-to-observe regions. Given a voxel normal $\mathbf{n}$, a local tangent frame $(\mathbf{t}_1,\mathbf{t}_2)$ is constructed. Directions are sampled by tilting the normal by angle $\beta$ and rotating around it by angle $\phi$, yielding $\mathbf{d}_{\text{tilt}} = \cos(\beta)\mathbf{n} + \sin(\beta)\left(\cos(\phi)\mathbf{t}_1+\sin(\phi)\mathbf{t}_2\right)$. A viewpoint is placed along each sampled direction as $\mathbf{p} = \mathbf{c}_{\text{voxel}} + h\,\mathbf{d}_{\text{tilt}}$, where $h$ is the same GSD-derived distance used for initial viewpoint generation. Candidates are validated using the same ray pre-filter, then merged into the viewpoint pool and incorporated into visibility and SBW updates.

\section{Selecting a Subset of Viewpoints}

Given the sensing-constrained candidate pool, the remaining
problem is to select a compact subset of viewpoints satisfying
the prescribed per-voxel $k$-coverage requirement. This induces
a set multi-cover formulation: let $V$ denote the surface voxel
set and each candidate $\mathbf{p}_j \in P$ correspond to a
subset $V_j \subseteq V$ of visible voxels. The objective is
to select $S \subseteq P$ such that each voxel is observed
at least $k$ times, when feasible, while minimizing $|S|$.

We propose a \textit{Globally Informed Greedy $k$-Coverage}
heuristic that dynamically prioritizes voxels with unmet
coverage demand and limited remaining visibility. For compact representation, let
$A \in \{0,1\}^{N \times M}$ denote the binary visibility
matrix, where $N=|V|$, $M=|P|$, and $A_{ij}=1$ if
viewpoint $\mathbf{p}_j$ observes voxel $v_i$, and $0$
otherwise.

\begin{algorithm}[H]
\small
    \caption{\textbf{Globally Informed Greedy $k$-Coverage}}
    \label{alg:algo2}
    \begin{algorithmic}[1]

    \Require Candidate viewpoints
    $P=\{\mathbf{p}_1,\dots,\mathbf{p}_M\}$,
    visibility matrix $A\in\{0,1\}^{N\times M}$,
    target coverage $k$

    \Ensure Selected viewpoint set $S\subseteq P$

    \State $S\gets\emptyset$
    \State $R\gets P$
    \Comment{Remaining viewpoints}

    \State $\mathbf{b}\gets A\mathbf{1}$
    \Comment{Initial SBW counts}
    \State $\mathbf{k}^*\gets
    \min(k\mathbf{1},\mathbf{b})$
    \State $\mathbf{cov}\gets\mathbf{0}$
    \State $\mathbf{b}_{rem}\gets\mathbf{b}$

    \While{$\exists\,i:\mathrm{cov}_i<k_i^*$}

        \State $\mathbf{r}\gets
        \max(\mathbf{0},\mathbf{k}^*-\mathbf{cov})$

        \State $\boldsymbol{\sigma}\gets
        \max(\mathbf{0},\mathbf{b}_{rem}-\mathbf{r})$

        \State $\boldsymbol{\omega}\gets
        \exp(-\boldsymbol{\sigma})
        \odot(\mathbf{r}/k)^{\odot 2}$

        \State $\mathbf{score}\gets
        A^\top\boldsymbol{\omega}$

        \State $\mathbf{p}^*\gets
        \arg\max_{\mathbf{p}_j\in R}
        \mathrm{score}_j$

        \If{no unmet voxel is seen by any $\mathbf{p}_j\in R$}
            \State \textbf{break}
        \EndIf

        \State $S\gets S\cup\{\mathbf{p}^*\}$
        \State $R\gets R\setminus\{\mathbf{p}^*\}$

        \State $\mathbf{cov}\gets
        \mathbf{cov}+A_{:,\mathbf{p}^*}$

        \State $\mathbf{b}_{rem}\gets
        \mathbf{b}_{rem}-A_{:,\mathbf{p}^*}$

    \EndWhile

    \State $\mathrm{changed}\gets\mathrm{True}$
    \Comment{Cleanup}

    \While{$\mathrm{changed}$}
        \State $\mathrm{changed}\gets\mathrm{False}$

        \For{each $\mathbf{p}\in S$}
            \If{removing $\mathbf{p}$ preserves
            $\mathbf{k}^*$-coverage}
                \State $S\gets S\setminus\{\mathbf{p}\}$
                \State $\mathrm{changed}\gets\mathrm{True}$
                \State \textbf{break}
            \EndIf
        \EndFor

    \EndWhile

    \State \Return $S$

    \end{algorithmic}
\end{algorithm}

Here, $\mathbf{b}$ represents the initial SBW counts,
$\mathbf{b}_{rem}$ represents the remaining candidate
visibility counts, and $\mathbf{cov}$ denotes the current
coverage counts. The operators $\min$, $\max$, $\exp$,
division, and exponentiation are applied elementwise;
$\odot$ denotes elementwise multiplication.
The notation $A_{:,\mathbf{p}^*}$ refers to the column
corresponding to the selected viewpoint.

Each voxel is assigned a dynamic weight that combines
a \emph{rarity} component, driven by the slack term
$\boldsymbol{\sigma}=
\max(\mathbf{0},\mathbf{b}_{rem}-\mathbf{r})$,
and an \emph{urgency} component, capturing the
remaining coverage deficit $\mathbf{r}$ relative to
the target $k$-coverage level.

As viewpoints are selected, the remaining visibility
counts are updated, so voxels with few remaining
viewpoint alternatives retain higher weights, while
those with many alternatives are down-weighted.
The urgency term emphasizes voxels with larger
remaining coverage deficits. Together, these terms
prioritize scarce and underserved surface regions.
The weighting function is heuristic and does not
establish a theoretical approximation guarantee.

The per-voxel target is capped as
$k^*(v)=\min(k,\mathrm{SBW}(v))$, since voxels
visible from fewer than $k$ candidate viewpoints
cannot satisfy the requested coverage requirement.
The augmentation stage aims to increase
$\mathrm{SBW}(v)$ so that $k^*(v)=k$ for all voxels.
When every surface voxel has sufficient candidate
visibility, the greedy selection can achieve full
$k$-coverage under the modeled visibility relation.
Otherwise, the reduced targets accommodate
limitations of the generated candidate set. A cleanup pass is applied to $S$ that iteratively
removes any selected viewpoint whose removal
preserves all per-voxel targets $k^*(v)$.

\section{Baseline Algorithms for Comparison}

We compare our method with a locally guided overlap-pruning heuristic and an ILP formulation that provides minimum-cardinality solutions to the discrete multi-cover problem when optimality is certified. As discussed in Section \ref{sec:related}, overlap-based pruning is widely used in coverage planning. We formalize this as \emph{Sequential Overlap-Based Viewpoint Pruning}, in which candidate viewpoints are processed sequentially and accepted only if their visible region contributes sufficient new coverage relative to previously selected viewpoints \cite{bircher2015structural,lee2022ceo}. While such heuristics promote redundancy, they do not explicitly enforce per-voxel multi-cover constraints. We therefore include this method as a representative redundancy-reduction baseline to assess how explicit $k$-coverage reasoning alters coverage behavior.

\begin{algorithm}[H]
\small
    \caption{\textbf{Sequential Overlap-Based Viewpoint Pruning}}
    \label{alg:approach1}
    \begin{algorithmic}[1]

    \Require Candidate viewpoints $P = \{\mathbf{p}_1, \dots, \mathbf{p}_M\}$, $V = \bigcup_{j=1}^{M} V_j$ (all visible voxels) where $V_j \subseteq V$ is the set of voxels visible from $\mathbf{p}_j$, overlap threshold $\tau \in [0,1]$
    \Ensure $S \subseteq P$

    \State $S \gets \emptyset$
    \State $G \gets \emptyset$ \Comment{{\footnotesize globally covered voxels}}
    \State Sort $P$ by chosen ordering (e.g.\ bottom-to-top by height)

    \For{each viewpoint $\mathbf{p}_j \in P$}
        \If{$S = \emptyset$} \Comment{{\footnotesize first viewpoint always accepted}}
            \State $S \gets S \cup \{\mathbf{p}_j\}$
            \State $G \gets G \cup V_j$
        \Else
            \State $\mathrm{overlap}_j \gets \dfrac{|V_j \cap G|}{|V_j|}$
            \If{$\mathrm{overlap}_j < \tau$} \Comment{{\footnotesize enough new coverage}}
                \State $S \gets S \cup \{\mathbf{p}_j\}$
                \State $G \gets G \cup V_j$
            \EndIf
        \EndIf
    \EndFor

    \State \Return $S$

    \end{algorithmic}
\end{algorithm}

The overlap threshold is defined as $\tau$ and measures overlap between each candidate viewpoint's view and the already covered voxels $G$. A viewpoint is selected only if its overlap with $G$ is strictly less than $\tau$. This ensures that views that mostly see already-covered voxels are pruned. We consider two sorting strategies as well. In vertical ordering, viewpoints are processed from bottom to top, and in visibility-based ordering, viewpoints are processed in descending order of the number of voxels they can observe.

This strategy exhibits several structural limitations. First, it introduces a \emph{starting-point bias}, as the first candidate is always accepted and early decisions disproportionately influence the final solution. This leads to \emph{order dependence}, where different candidate orderings can produce substantially different viewpoint sets. More fundamentally, overlap pruning provides \emph{no explicit coverage guarantee}: unfavorable thresholds or orderings may leave voxels uncovered even when feasible solutions exist. The selection process is also local in nature, as each candidate is evaluated only with respect to previously selected viewpoints rather than the global coverage state. Decisions are therefore made sequentially based on current overlap information without explicitly reasoning about unmet coverage requirements across the entire voxel set.

As a global baseline, we formulate the $k$-coverage requirement as an ILP, yielding an exact combinatorial model of the set multi-cover problem that minimizes the number of selected viewpoints subject to explicit per-voxel coverage constraints. The optimization is performed over the generated candidate set, and a minimum-cardinality solution is established when optimality is certified.

\begin{algorithm}[H]
\small
    \caption{\textbf{ILP-Based Minimum-Cardinality Viewpoint Selection}}
    \label{alg:approach10}
    \begin{algorithmic}[1]

    \Require Candidate viewpoints $P = \{\mathbf{p}_1, \dots, \mathbf{p}_M\}$, $V = \bigcup_{j=1}^{M} V_j$ (all visible voxels) where $V_j \subseteq V$ is the set of voxels visible from $\mathbf{p}_j$, target coverage $k$
    \Ensure Selected viewpoint set $S \subseteq P$

    \For{each voxel $v \in V$}
        \State $k^*(v) \gets \min\bigl(k,\; \mathrm{SBW}(v)\bigr)$ \Comment{per-voxel coverage target}
    \EndFor

    \State Define binary variables $x_j \in \{0,1\}$ for each $\mathbf{p}_j \in P$

    \State \textbf{Solve:}
    \Statex \quad Minimize $\displaystyle\sum_{j=1}^{M} x_j$
    \Statex \quad subject to $\displaystyle\sum_{j \,:\, v \in V_j} x_j \;\geq\; k^*(v) \quad \forall\, v \in V$

    \State $S \gets \{\mathbf{p}_j \in P : x_j = 1\}$

    \State \Return $S$

    \end{algorithmic}
\end{algorithm}

\noindent
The ILP is solved using a branch-and-bound procedure, in which integrality constraints are relaxed to obtain linear-programming lower bounds, and subproblems are pruned when they cannot improve the best feasible integer solution. If optimality is certified within the imposed time limit, the solution is provably optimal over the generated candidate set; otherwise, the best feasible solution obtained is returned.

Although the ILP yields a minimum-cardinality solution when optimality is certified, its worst-case complexity is exponential due to the NP-hardness of set multi-cover. The formulation introduces one binary decision variable per candidate viewpoint and one coverage constraint per voxel, so the instance size grows quickly with finer voxelizations and richer augmentation. In practice, solve time depends not only on the number of variables/constraints but also on the structure of the visibility matrix (e.g., redundancy and overlap patterns), and some large instances may not certify optimality within a fixed time budget. We therefore use the ILP as a reference for comparing viewpoint-set cardinality, with certified optimal solutions providing a minimum-cardinality benchmark over the generated candidate set.

\section{Experiments}
\label{sec:experiments}

\begin{figure}[H]
\centering
\begin{minipage}{0.24\columnwidth}
\centering
\includegraphics[width=\linewidth]{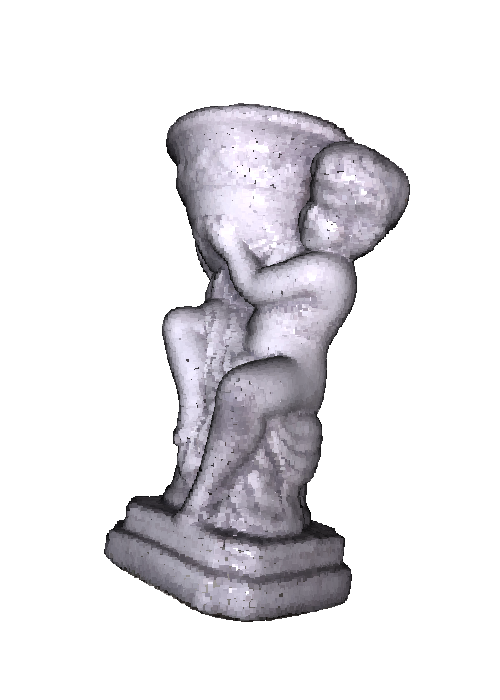}
{\scriptsize (a) Angel Statue (Point Cloud)}
\end{minipage}\hfill
\begin{minipage}{0.24\columnwidth}
\centering
\includegraphics[width=\linewidth]{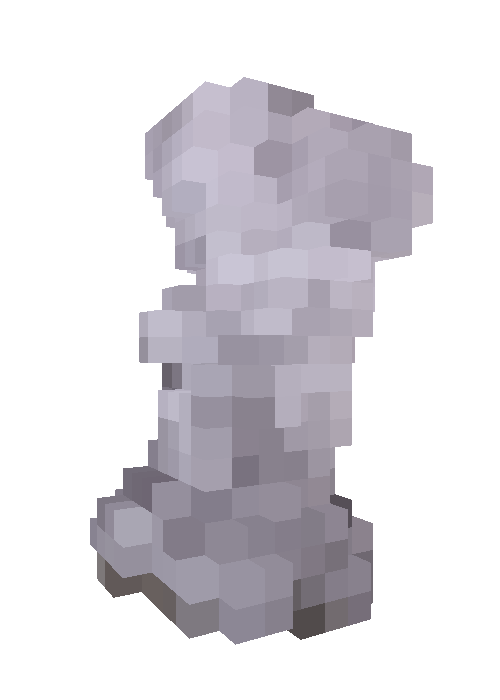}
{\scriptsize (b) Angel Statue (Voxelized)}
\end{minipage}\hfill
\begin{minipage}{0.24\columnwidth}
\centering
\includegraphics[width=\linewidth]{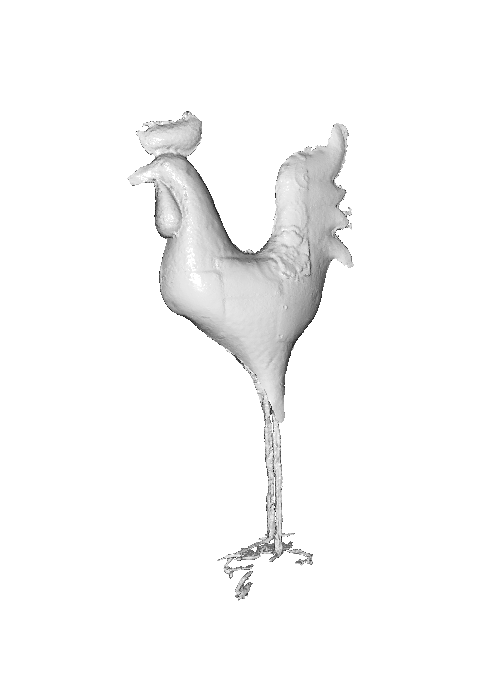}
{\scriptsize (c) Rooster (Mesh)}
\end{minipage}\hfill
\begin{minipage}{0.24\columnwidth}
\centering
\includegraphics[width=\linewidth]{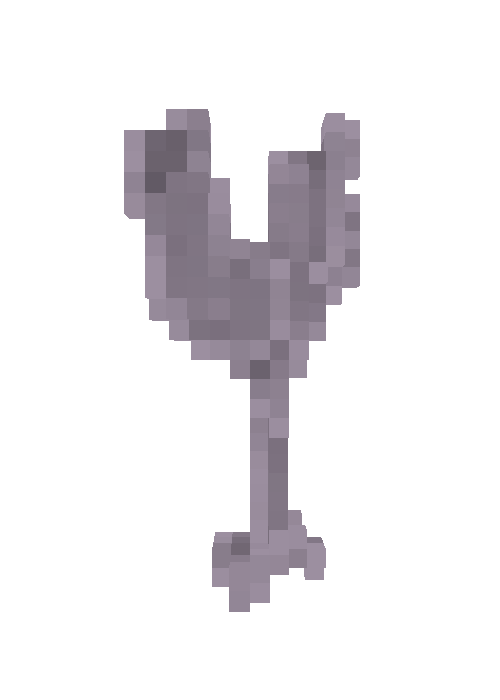}
{\scriptsize (d) Rooster (Voxelized)}
\end{minipage}
\caption{Evaluation structures with voxelized representations}
\label{fig:structures}
\end{figure}

The proposed pipeline is evaluated on two publicly accessible 3D models \cite{python3d_analysis_libraries} with distinct geometric characteristics, shown in Fig.~\ref{fig:structures}. Experimental parameters are summarized in Table~\ref{tab:exp_config}. The selected voxel sizes span fine-to-coarse abstractions. Overlap thresholds reflect redundancy regimes common in photogrammetric practice. Visibility augmentation parameters sweep the specified tilt and azimuth ranges to generate additional candidates for geometrically challenging regions where normal-aligned viewpoints alone may provide insufficient candidate visibility for the requested $k$-coverage. The complete pipeline was implemented in Python using the Open3D library.

\begin{table}[H]
\centering
\caption{Experimental parameters.}
\label{tab:exp_config}

\setlength{\tabcolsep}{1pt}
\renewcommand{\arraystretch}{1.12}

\begin{tabular}{ll}
\toprule
\textbf{Parameter} & \textbf{Values} \\
\midrule

Voxel size ($s$)
& $\{0.04,\,0.06,\,0.08,\,0.10\}\ \mathrm{m}$ \\

GSD$_{\max}$
& $0.5 \mathrm{mm/pixel}$ \\

GSD setpoint
& $0.25 \mathrm{mm/pixel}$ \\

$\alpha$
& $45^\circ$ \\

$R_s$
& $0.3394\ \mathrm{m}$ \\

$h$
& $0.2400\ \mathrm{m}$ \\

Overlap threshold
& $\{50\%,\,70\%\}$ \\

Sorting Criteria
& \{Vertical, Visibility\} \\

Coverage requirement ($k$)
& $\{1,\,3,\,5\}$ \\

Tilt angle ($\beta$) (Visibility augmentation)
& $\{1^\circ - 90^\circ\}$ (step $1^\circ$) \\

Azimuth rotation ($\phi$) (Visibility augmentation)
& $\{0^\circ - 360^\circ\}$ (step $30^\circ$) \\

Time limit for ILP solution
& $3600\ \mathrm{s}$ \\

\bottomrule
\end{tabular}
\end{table}

Sequential Overlap-Based Viewpoint Pruning is evaluated only on a voxel resolution of 0.10 m to illustrate its behavior under different overlap percentages and ordering strategies. In contrast, Globally Informed Greedy $k$-Coverage and ILP-Based Minimum-Cardinality Viewpoint Selection are evaluated across all voxel sizes and $k$ values.

\section{Results}

\begin{figure}[H]
    \centering
    \includegraphics[width=0.6\linewidth]{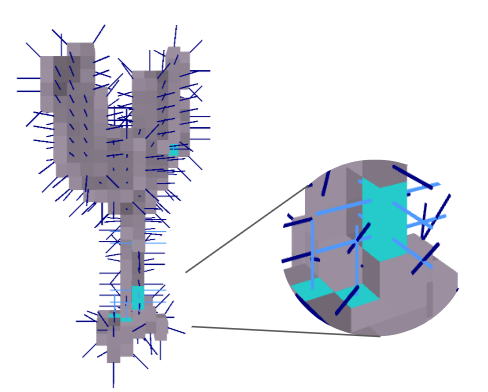}
    \caption{\legendbox{multi_norm_voxel} Multi-normal voxels and 
\legendbox{multi_norm} their retained outward-facing normals.}
    \label{fig:rooster}
\end{figure}

Table~\ref{tab:voxel_stats} summarizes voxel and viewpoint statistics. The total occupied voxels (TV) are separated into internal voxels (IV) and surface voxels (SV), with internal voxels excluded from candidate generation. For the Angel Statue, the initial waypoint (IW) count matches the surface voxel count, indicating one initial viewing direction per surface voxel. In contrast, the Rooster produces more initial viewpoints than surface voxels because some voxels retain multiple outward-facing normals (Fig.~\ref{fig:rooster}). The blocked-viewpoint (BW) counts report candidates removed by the ray-based occlusion pre-filter.

\begin{table}[H]
\centering
\caption{\small Voxelization statistics across resolutions for both structures}
\label{tab:voxel_stats}

{\scriptsize
\begin{tabular}{lcccccccc}
\toprule
 & \multicolumn{4}{c}{\textbf{Angel Statue}} 
 & \multicolumn{4}{c}{\textbf{Rooster}} \\
\cmidrule(lr){2-5} \cmidrule(lr){6-9}
\textbf{VS} 
& 0.04 & 0.06 & 0.08 & 0.10 
& 0.04 & 0.06 & 0.08 & 0.10 \\
\midrule

\textbf{TV} 
& 4178 & 1905 & 1064 & 695 
& 2131 & 952 & 536 & 332 \\

\textbf{IV}
& 1397 & 653 & 350 & 222 
& 599 & 217 & 114 & 49 \\

\textbf{SV}
& 2781 & 1252 & 714 & 473 
& 1532 & 735 & 422 & 283 \\

\textbf{IW}
& 2781 & 1252 & 714 & 473 
& 1540 & 741 & 435 & 294 \\

\textbf{BW} 
& 67 & 20 & 7 & 4 
& 47 & 27 & 9 & 4 \\
\bottomrule
\end{tabular}
}

\end{table}

Table~\ref{tab:visibility_stats} summarizes candidate visibility before and after targeted augmentation. Here, not covered (NC) denotes voxels with no visible candidate viewpoint and partially covered (PC) denotes voxels with fewer than $k$ visible candidates; NC$^*$ and PC$^*$ are the corresponding counts after augmentation. Across both structures and the evaluated voxel sizes and coverage levels, augmentation substantially reduces these deficiencies and establishes sufficient candidate visibility for the requested $k$ in most configurations. A small number of voxels remain undercovered in some cases. The additional viewpoint (AW) counts generally increase with the requested coverage level and vary with voxel resolution and geometry, reaching large candidate pools in several configurations. These results characterize feasibility of the generated candidate set before subset selection; they do not themselves measure coverage achieved by the final selected viewpoints.

To assess compliance with the adopted sensing model, we evaluate GSD for each retained viewpoint--voxel visibility pair, using the modeled voxel-center observation geometry. Table~\ref{tab:gsd_values} reports the minimum, maximum, mean, and standard deviation for each configuration. All reported maxima are at or below the prescribed $\mathrm{GSD}_{\max}=0.5\,\mathrm{mm/px}$. Thus, the retained visibility pairs satisfy the evaluated GSD bound under the geometric model; these statistics do not establish image quality over entire voxel surfaces or downstream reconstruction accuracy.

\begin{table}[H]
\centering
\caption{Visibility Statistics Before and After Viewpoint Augmentation}
\label{tab:visibility_stats}

{\scriptsize
\setlength{\tabcolsep}{1.9pt}
\renewcommand{\arraystretch}{1.15}

\begin{tabular}{ccrrrrrrrrrr}
\toprule
 &  & \multicolumn{5}{c}{\textbf{Angel Statue}} 
 & \multicolumn{5}{c}{\textbf{Rooster}} \\
\cmidrule(lr){3-7} \cmidrule(lr){8-12}

\textbf{$k$} & VS
& NC & PC & AW & NC$^*$ & PC$^*$
& NC & PC & AW & NC$^*$ & PC$^*$ \\
\midrule

\multirow{4}{*}{1}
& 0.04 & 5 & 0 & 291 & 1 & 0 & 2 & 0 & 635 & 1 & 0 \\
& 0.06 & 2 & 0 & 755 & 0 & 0 & 8 & 0 & 3490 & 0 & 0 \\
& 0.08 & 0 & 0 & 0   & 0 & 0 & 5 & 0 & 2004 & 0 & 0 \\
& 0.10 & 2 & 0 & 500 & 0 & 0 & 3 & 0 & 994  & 0 & 0 \\
\midrule

\multirow{4}{*}{3}
& 0.04 & 5 & 9  & 1797  & 1 & 1 & 2 & 18 & 8992  & 1 & 0 \\
& 0.06 & 2 & 9  & 4082  & 0 & 0 & 8 & 30 & 23887 & 0 & 0 \\
& 0.08 & 0 & 5  & 2501  & 0 & 0 & 5 & 35 & 27426 & 0 & 0 \\
& 0.10 & 2 & 12 & 10386 & 0 & 0 & 3 & 31 & 26469 & 0 & 0 \\
\midrule

\multirow{4}{*}{5}
& 0.04 & 5 & 20  & 5539  & 0 & 1 & 2 & 61 & 36155 & 0 & 0 \\
& 0.06 & 2 & 26  & 11744 & 0 & 0 & 8 & 75 & 59651 & 0 & 0 \\
& 0.08 & 0 & 47  & 30450 & 0 & 0 & 5 & 78 & 63369 & 0 & 0 \\
& 0.10 & 2 & 109 & 84351 & 0 & 0 & 3 & 86 & 69633 & 0 & 0 \\
\bottomrule
\end{tabular}
}
\end{table}

\begin{table}[H]
\centering
\caption{GSD statistics over viewpoint–voxel pairs (all values are reported in mm/px, Std = standard deviation).}
\label{tab:gsd_values}

{\scriptsize
\setlength{\tabcolsep}{2.4pt}
\renewcommand{\arraystretch}{1.15}

\begin{tabular}{ccrrrrrrrr}
\toprule
\multirow{2}{*}{\textbf{$k$}} & \multirow{2}{*}{\textbf{VS}}
& \multicolumn{4}{c}{\textbf{Angel Statue}}
& \multicolumn{4}{c}{\textbf{Rooster}} \\
\cmidrule(lr){3-6} \cmidrule(lr){7-10}
& & \textbf{Min} & \textbf{Max} & \textbf{Mean} & \textbf{Std}
  & \textbf{Min} & \textbf{Max} & \textbf{Mean} & \textbf{Std} \\
\midrule

\multirow{4}{*}{1}
& 0.04 & 0.102 & 0.500 & 0.330 & 0.069 & 0.083 & 0.500 & 0.331 & 0.071 \\
& 0.06 & 0.118 & 0.500 & 0.335 & 0.064 & 0.105 & 0.500 & 0.325 & 0.069 \\
& 0.08 & 0.167 & 0.500 & 0.335 & 0.072 & 0.082 & 0.500 & 0.326 & 0.063 \\
& 0.10 & 0.150 & 0.495 & 0.328 & 0.080 & 0.174 & 0.492 & 0.330 & 0.067 \\
\midrule

\multirow{4}{*}{3}
& 0.04 & 0.102 & 0.500 & 0.325 & 0.069 & 0.082 & 0.500 & 0.321 & 0.069 \\
& 0.06 & 0.118 & 0.500 & 0.327 & 0.065 & 0.097 & 0.500 & 0.317 & 0.064 \\
& 0.08 & 0.166 & 0.500 & 0.325 & 0.069 & 0.081 & 0.500 & 0.313 & 0.061 \\
& 0.10 & 0.150 & 0.499 & 0.321 & 0.068 & 0.135 & 0.496 & 0.314 & 0.062 \\
\midrule

\multirow{4}{*}{5}
& 0.04 & 0.088 & 0.500 & 0.324 & 0.066 & 0.050 & 0.500 & 0.313 & 0.068 \\
& 0.06 & 0.102 & 0.500 & 0.321 & 0.065 & 0.097 & 0.500 & 0.317 & 0.063 \\
& 0.08 & 0.117 & 0.500 & 0.321 & 0.062 & 0.081 & 0.500 & 0.315 & 0.060 \\
& 0.10 & 0.121 & 0.499 & 0.319 & 0.062 & 0.111 & 0.499 & 0.314 & 0.062 \\
\bottomrule
\end{tabular}
}
\end{table}

Table~\ref{tab:Sequential} reports the original evaluation's full-coverage (FC), partial-coverage (PC), and not-covered (NC) categories for Sequential Overlap-Based Viewpoint Pruning. These reported categories describe coverage outcomes after pruning, whereas the overlap threshold determines whether an additional candidate is accepted. Increasing the overlap threshold from $50\%$ to $70\%$ generally increases the number of selected viewpoints (TS) and reduces uncovered regions. Nevertheless, the reported full-coverage category does not reach $100\%$ in any tested configuration. The results also show sensitivity to ordering: vertical and visibility-based sorting produce substantially different viewpoint counts and coverage distributions under the same threshold.

\begin{table}[H]
\centering
\caption{Sequential Overlap-Based Viewpoint Pruning results at voxel size 0.10 m}
\label{tab:Sequential}

{\scriptsize
\setlength{\tabcolsep}{3pt}
\renewcommand{\arraystretch}{1.12}

\begin{tabular}{cc*{8}{r}}
\toprule
O & S
& \multicolumn{4}{c}{\textbf{Angel Statue}}
& \multicolumn{4}{c}{\textbf{Rooster}} \\
\cmidrule(lr){3-6} \cmidrule(lr){7-10}
&
& TS & FC (\%) & PC (\%) & NC (\%)
& TS & FC (\%) & PC (\%) & NC (\%) \\
\midrule

\multirow{2}{*}{50}
& Vertical
& 171 & 32.14 & 63.00 & 4.86
& 135 & 32.99 & 64.29 & 2.72 \\

& Visibility
& 71 & 21.99 & 75.90 & 2.11
& 56 & 27.89 & 69.73 & 2.38 \\

\midrule

\multirow{2}{*}{70}
& Vertical
& 253 & 32.98 & 66.60 & 0.42
& 194 & 39.12 & 60.54 & 0.34 \\

& Visibility
& 80 & 13.32 & 86.26 & 0.42
& 70 & 19.05 & 80.27 & 0.68 \\

\bottomrule
\end{tabular}
}

\end{table}

\begin{table*}[t]
\centering
\caption{Comparison between Globally Informed Greedy $k$-Coverage and ILP-Based Minimum-Cardinality Viewpoint Selection across voxel sizes and $k$-coverage levels. 
Difference (\%) is computed relative to the reported ILP viewpoint count; Opt. indicates whether ILP optimality was certified.}
\label{tab:comparision_greedy_ilp}

\setlength{\tabcolsep}{2.6pt}
\renewcommand{\arraystretch}{1.12}

\begin{tabular}{cc rrrr rrrr rrrrc rrrrc rr}
\toprule

& &
\multicolumn{8}{c}{\textbf{Globally Informed Greedy $k$-Coverage}} &
\multicolumn{10}{c}{\textbf{ILP-Based Viewpoint Selection}} &
\multicolumn{2}{c}{\textbf{Difference of WS (\%)}} \\
\cmidrule(lr){3-10} \cmidrule(lr){11-20} \cmidrule(lr){21-22}

& &
\multicolumn{4}{c}{\textbf{Angel Statue}} &
\multicolumn{4}{c}{\textbf{Rooster}} &
\multicolumn{5}{c}{\textbf{Angel Statue}} &
\multicolumn{5}{c}{\textbf{Rooster}} &
\textbf{Angel Statue} &
\textbf{Rooster} \\

\cmidrule(lr){3-6}
\cmidrule(lr){7-10}
\cmidrule(lr){11-15}
\cmidrule(lr){16-20}
\cmidrule(lr){21-22}

\textbf{$k$} & \textbf{VS} &
\textbf{WA} & \textbf{WS} & \textbf{CR} & \textbf{FC(\%)} &
\textbf{WA} & \textbf{WS} & \textbf{CR} & \textbf{FC(\%)} &
\textbf{WA} & \textbf{WS} & \textbf{CR} & \textbf{FC(\%)} & \textbf{Opt.} &
\textbf{WA} & \textbf{WS} & \textbf{CR} & \textbf{FC(\%)} & \textbf{Opt.} &
\textbf{} & \textbf{} \\

\midrule

\multirow{4}{*}{1}
& 0.04 & 3005 & 142 & 5 & 99.96  & 2128 & 110 & 3 & 99.93  & 3005 & 122 & 0 & 99.96 & False & 2128 & 95 & 0 & 99.93 & True  & 16.39 & 15.79 \\
& 0.06 & 1987 & 111 & 0 & 100.00 & 4204 & 85  & 1 & 100.00 & 1987 & 96  & 0 & 100.00 & True  & 4204 & 76 & 0 & 100.00 & True  & 15.62 & 11.84 \\
& 0.08 & 707  & 92  & 1 & 100.00 & 2430 & 78  & 1 & 100.00 & 707  & 87  & 0 & 100.00 & True  & 2430 & 73 & 0 & 100.00 & True  & 5.75  & 6.85  \\
& 0.10 & 969  & 81  & 0 & 100.00 & 1284 & 69  & 0 & 100.00 & 969  & 76  & 0 & 100.00 & True  & 1284 & 64 & 0 & 100.00 & True  & 6.58  & 7.81  \\
\midrule

\multirow{4}{*}{3}
& 0.04 & 4511 & 382 & 8 & 99.93  & 10485 & 283 & 8 & 99.93  & 4511 & 336 & 0 & 99.93  & False & 10485 & 248 & 0 & 99.93 & False & 13.69 & 14.11 \\
& 0.06 & 5314 & 307 & 6 & 100.00 & 24601 & 239 & 2 & 100.00 & 5314 & 275 & 0 & 100.00 & True  & 24601 & 209 & 0 & 100.00 & True  & 11.64 & 14.35 \\
& 0.08 & 3208 & 275 & 0 & 100.00 & 27852 & 206 & 3 & 100.00 & 3208 & 267 & 0 & 100.00 & True  & 27852 & 189 & 0 & 100.00 & True  & 3.00  & 8.99  \\
& 0.10 & 10855& 239 & 4 & 100.00 & 26759 & 175 & 2 & 100.00 & 10855& 229 & 0 & 100.00 & True  & 26759 & 165 & 0 & 100.00 & True  & 4.37  & 6.06  \\
\midrule

\multirow{4}{*}{5}
& 0.04 & 8253 & 613 & 17 & 99.96  & 37648 & 441 & 12 & 100.00 & 8253 & 543 & 0 & 99.96  & False & 37648 & 390 & 0 & 100.00 & False & 12.89 & 13.08 \\
& 0.06 & 12976& 504 & 9  & 100.00 & 60365 & 378 & 8  & 100.00 & 12976& 456 & 0 & 100.00 & True  & 60365 & 336 & 0 & 100.00 & True  & 10.53 & 12.50 \\
& 0.08 & 31157& 426 & 9  & 100.00 & 63795 & 334 & 7  & 100.00 & 31157& 383 & 0 & 100.00 & True  & 63795 & 292 & 0 & 100.00 & True  & 11.23 & 14.38 \\
& 0.10 & 84820& 343 & 4  & 100.00 & 69923 & 279 & 3  & 100.00 & 84820& 301 & 0 & 100.00 & True  & 69923 & 246 & 0 & 100.00 & True  & 13.95 & 13.41 \\
\bottomrule

\end{tabular}
\end{table*}

In contrast to the overlap-pruning baseline, the globally informed greedy method explicitly tracks outstanding per-voxel coverage requirements and reports full or near-full requested $k$-coverage across the evaluated configurations (Table~\ref{tab:comparision_greedy_ilp}). This comparison illustrates the difference between pruning on accumulated overlap and selecting viewpoints to satisfy explicit coverage demands.

Table~\ref{tab:comparision_greedy_ilp} compares the greedy and ILP approaches over the same generated candidate pools. It reports available viewpoints (WA), selected viewpoints (WS), viewpoints removed during cleanup (CR), reported full coverage (FC), and ILP optimality certification (Opt.), along with the relative difference in selected viewpoint counts. In this table, FC refers to attainment of the requested $k$ observations; cases with fewer than $k$ admissible candidates for some voxels cannot attain complete requested $k$-coverage, even though both selection algorithms use capped targets for those voxels.

The cleanup stage removes a small number of viewpoints from the greedy solutions without changing satisfaction of their capped per-voxel targets. No viewpoints are removed from the reported ILP solutions during cleanup. This is consistent with minimality where ILP optimality was certified, but does not independently establish optimality for the non-certified cases.

When the generated pool contains at least $k$ admissible viewpoints for each target voxel, both methods report full requested $k$-coverage. The few configurations with slightly lower FC values (99.93--99.96\%) coincide with residual candidate-visibility deficiencies reported in Table~\ref{tab:visibility_stats}. Such configurations satisfy the capped targets where possible but do not establish unconditional $k$-coverage over all surface voxels.

The ILP provides a minimum-cardinality benchmark for the discrete candidate pool only when the solver certifies optimality. Otherwise, its reported viewpoint count is a feasible incumbent rather than a proven optimum. The greedy solutions use 3.00--16.39\% more viewpoints than the corresponding reported ILP solutions across the evaluated configurations. This is a comparison of selected-set cardinality, not a certified optimality gap for every case. The results show that the interpretable rarity- and deficit-weighted selection rule produces compact viewpoint sets with counts comparable to those returned by the ILP solver, although no computational speedup or approximation bound is established by these experiments.

\section{Conclusion}
This work formulates aerial inspection viewpoint selection as a GSD-constrained $k$-coverage problem over voxelized surface representations. Candidate viewpoints are constructed and filtered under a geometric sensing model, while targeted augmentation increases the availability of candidate observations for undercovered voxels. Set multi-cover optimization then enables explicit evaluation of per-voxel coverage requirements over the resulting discrete candidate pool.

The proposed Globally Informed Greedy $k$-Coverage heuristic prioritizes remaining coverage deficits and scarce viewing alternatives through an interpretable, dynamically weighted selection rule. Across the evaluated structures, voxel sizes, and coverage levels, it achieves full or near-full requested $k$-coverage with compact viewpoint sets, requiring 3.00--16.39\% more viewpoints than the corresponding ILP solutions. The ILP comparison provides a discrete minimum-cardinality reference where optimality is certified; neither global camera-pose optimality nor a theoretical approximation guarantee is claimed.

The present study assumes known, static structure geometry and evaluates coverage using modeled voxel-center sensing and visibility constraints. It does not assess minimum viewpoint separation, UAV reachability or dynamics, collision-safe trajectories, pipeline runtime, or downstream reconstruction quality; the evaluation is also limited to two structures. Future work may incorporate execution feasibility and reconstruction-based assessment while retaining explicit coverage reasoning.

\section*{ACKNOWLEDGMENT}
Generative AI tools were used to assist in limited aspects of manuscript preparation and software development. The Cursor IDE was used to support pipeline implementation and visualization generation. ChatGPT (OpenAI) and Claude (Anthropic) assisted in improving manuscript clarity, refining technical explanations, and organizing the literature review. All algorithmic formulations, experimental design, analyses, and final written content were developed, verified, and validated by the authors.

\bibliographystyle{IEEEtran}
\bibliography{references}

\end{document}